\documentclass{article}

\usepackage[preprint]{aaj2026}
\usepackage{natbib}

\usepackage[utf8]{inputenc} 
\usepackage[T1]{fontenc}    
\usepackage{microtype}      
\usepackage{hyperref}       
\usepackage{url}            
\usepackage{booktabs}       
\usepackage{amsfonts}       
\usepackage{nicefrac}       
\usepackage{xcolor}         
\usepackage{amsmath}
\usepackage{graphicx}
\usepackage{textgreek}
\usepackage{tabularx}
\usepackage{multirow}
\usepackage{todonotes}
\usepackage{tikz}
\usepackage{subcaption}
\usepackage{CJKutf8}
\usepackage{threeparttable}
\usepackage{subcaption}
\usepackage{rotating}

\newcommand{\keywords}[1]{%
  \par\vspace{0.5em}\noindent\textbf{Keywords:} #1\par\vspace{0.5em}%
}
\usepackage{tabularx}

\title{Chinese sensorimotor and embodiment norms for 3,000 lexicalized concepts}

\author{%
  Jing Chen \\
  Department of Psychology \\
  University of Milano-Bicocca \\
  Department of Chinese and Bilingual Studies \\
  The Hong Kong Polytechnic University \\
  \texttt{jing.chen@unimib.it} \\
  \And
  Gábor Parti \\
  Department of Chinese and Bilingual Studies \\
  The Hong Kong Polytechnic University \\
  \texttt{gabor.parti@connect.polyu.hk} \\
  \And
  Yin Zhong \\
  Center for Language Education \\
  The Hong Kong University of Science and Technology \\
  \texttt{lcyinzhong@ust.hk} \\
  \AND
  Chu-Ren Huang \\
  Department of Chinese and Bilingual Studies \\
  The Hong Kong Polytechnic University \\
  \texttt{churen.huang@unimib.it} \\
  \And
  Marco Marelli \\
  Department of Psychology \\
  University of Milano-Bicocca \\
  \texttt{marco.marelli@unimib.it} \\
}

\begin{document}
\begin{CJK}{UTF8}{gkai}

\maketitle

\begin{abstract}
Understanding how conceptual knowledge is grounded in bodily experience, and to what extent machine systems can acquire such knowledge without direct sensorimotor experience, are central questions in both cognitive science and embodied artificial intelligence research. Large-scale normative resources are essential for investigating these questions empirically, yet such resources remain sparse for non-Indo-European languages. We present a novel normative database for 3,000 lexicalized concepts in Mandarin Chinese, comprising 11-dimensional sensorimotor ratings and unidimensional embodiment ratings collected from 378 native Mandarin speakers. The ratings demonstrate high reliability and strong cross-norm validity with existing Chinese resources, each of which covers fewer words and a subset of the 11 sensorimotor dimensions. In a validation study, we tested new variables derived from a theoretically motivated metric, \textit{Perceptual Strength of Embodiment} (PSE) \citep{huang2025linguistic}, together with seven common composite variables, on lexical decision tasks. The results suggest that PSE-Sensorimotor and \textit{Minkowski-3} are the strongest composite predictors of lexical decision performance, capturing the facilitatory effects of sensorimotor information on lexical processing. A further exploratory study showed that sensorimotor ratings are substantially recoverable from purely linguistic representations using simple regression models (mean Spearman $\rho = .62$ across dimensions), though recovery varied markedly: visual and auditory dimensions yielded higher correspondence than chemosensory ones. Representational similarity analysis further showed that the relational geometry of the sensorimotor space is also partially recoverable ($\rho = .540$), consistent with the view that distributional language use encodes aspects of embodied conceptual structure. The norms and code are available at \url{https://osf.io/h4pv2/overview?view_only=ab8d977c2d08464586a605401408ec76}.
\end{abstract}

\keywords{Embodied cognition, Sensorimotor strength, Embodiment, Linguistic representation, Mandarin Chinese}

How do humans mentally represent concepts? An emerging view in cognitive science holds that humans exploit multiple representation formats that collectively shape conceptual knowledge and support their interactions with the environment (\citealp[e.g.,][]{kaup2025interplay, kaup2024modal}). Among these, grounded representation may be considered a privileged characteristic of biological systems: it posits that mental representations are reactivations of prior sensorimotor experiences of referents in the physical world \citep{barsalou1999perceptual, wilson2002six, vigliocco2009toward, lynott2020lancaster}. Imagine encountering the concept of \textit{jiǎozi} (饺子, \textit{Chinese dumplings}) for the first time: we acquire its representation through direct bodily experience, taking in how they look, smell, and taste, along with the motor actions involved in preparing and eating them.

Sensorimotor information plays a well-established role in conceptual processing, as evidenced by converging behavioral and neuroimaging research \citep[e.g.,][]{kiefer2012conceptual, korner2023embodied, wellsby2014developing}. Behaviorally, perceptual information captures a unique dimension of word meaning and accounts for substantial variance in lexical decision performance \citep{juhasz2011tangible, juhasz2013sensory}, and explains more variance than other dimensions such as \textit{concreteness} or \textit{imageability} \citep{connell2012strength}. At the neural level, semantic decisions activate modality-specific cortical areas \citep[e.g.,][]{binder2011neurobiology}: \textit{tactile} knowledge is associated with somatosensory and motor activation \citep{goldberg2006perceptual}, and information from these unimodal regions converges into multimodal integration zones \citep{fernandino2016heteromodal,fernandino2016concept, tong2022distributed}, consistent with a distributed encoding of conceptual knowledge. 

This body of evidence motivates the development of large-scale normative databases providing sensorimotor information, though the specific dimensions vary across studies. For example, \citet{juhasz2011tangible, juhasz2013sensory} collected modality-general sensory experience ratings (SERs), which ask participants to rate the degree to which a word evokes a sensory experience without specifying a particular modality. Subsequent studies adopted a modality-specific approach, asking participants to rate the degree to which a concept is experienced through each of the five senses separately \citep[e.g.,][]{vergallito2020perceptual, lynott2009modality, lynott2013modality, chen2019mandarin, speed2017dutch}, which is generally considered more informative as it better captures the multimodal nature of conceptual representations \citep{lynott2013modality} and the systematic yet implicit mappings between orthographic and phonological forms and sensorimotor information \citep[e.g.,][]{chen2019mandarin, lynott2013modality, loca2025can}. In addition, interoception, referring to sensations arising from internal bodily states, was proposed as a sixth perceptual modality, and argued to play a particularly prominent role in representing abstract concepts such as emotion words \citep{connell2018interoception}. 

Beyond perceptual modalities, a growing body of evidence points to the importance of motor content in conceptual processing \citep[e.g.,][]{hauk2004somatotopic, boulenger2009grasping, ruschemeyer2007comprehending}. \citet{diez2018normative} collected norms for 750 Spanish object concepts that included ratings on motor attributes such as motoricity and graspability. Building on this, \citet{san2020motor} introduced motor content norms for 4,565 Spanish verbs, rating the degree to which different body parts (i.e., fingers, hands, legs, etc.) are involved in the movements associated with each concept. Building further on body-part specificity, \citet{lynott2020lancaster} collected action ratings across five effectors (i.e., hand/arm, foot/leg, head, mouth, and torso) for approximately 40,000 concepts. Together with ratings on six perceptual modalities, the resulting 11-dimensional Lancaster sensorimotor norms represent the most comprehensive resource of this kind to date, and have since been extended to other languages, including Italian \citep{repetto2023italian} and Chinese \citep{zhong2022sensorimotor}.

The multidimensional nature of sensorimotor information has been considered a bottleneck for large language models, i.e., artificial systems without grounding capacities, in acquiring rich conceptual representations \citep[e.g.,][]{xu2025large, ciaccio2025refining, lin2025six}. Developing normative resources for sensorimotor information is therefore crucial, both for understanding the regularities among different sensorimotor dimensions and as training data for improving model performance on tasks requiring grounding information.

In this study, we introduce a novel sensorimotor norming database for Mandarin Chinese, a language that has lacked large-scale comprehensive resources of this kind, though it is the most extensive non-English language in the training data for the current generation of large language models \citep[e.g.,][]{lu2025cultural}. Existing sensorimotor resources in Mandarin Chinese remain limited in scope and dimensionality. For example, \citet{chen2019mandarin} provided five perceptual modalities for 232 words, \citet{wu2024sensory} collected modality-general SERs (one dimensional rating) for 1,130 words and \citet{yi2025perception} extended this approach to 4,923 Chinese words translated from an English dataset \citep{warriner2013norms}, and \citet{britton2024embodiment} collected embodiment ratings for 689 words measuring how strongly each word evokes bodily experience. \citet{zhong2022sensorimotor} introduced the first 11-dimensional sensorimotor norms following the Lancaster framework \citep{lynott2020lancaster} but restricted to 664 nouns, limiting its applicability across broader lexical and grammatical contexts. 

To address these gaps, we collected 11-dimensional sensorimotor ratings, together with unidimensional embodiment ratings, for 3,000 frequently used lexicalized concepts in Mandarin Chinese (perceptual: visual, auditory, gustatory, olfactory, haptic, and interoception; action: leg/foot, hand/arm, mouth/throat, head, and torso), and reported descriptive statistics and correlation analyses in Study 1. In Study 2, we validated the database on lexical decision data and found facilitatory effects of sensorimotor information on lexical processing, particularly identifying a newly proposed metric as the best predictor \citep{huang2025linguistic}. In Study 3, we examined to what extent linguistic representations can predict sensorimotor ratings through simple regression analyses \citep{chersoni-etal-2020-automatic} to echo the debate on symbolic grounding.



\section{Study 1: Sensorimotor and embodiment ratings for 3,000 Chinese lexicalized concepts}
\label{sec:study1}

In this section, we present sensorimotor ratings for 3,000 Mandarin Chinese lexicalized concepts collected from native speakers, primarily following the established paradigm (e.g., Lancaster sensorimotor norms \citealt{lynott2020lancaster}; Chinese nominal norms \citealt{zhong2022sensorimotor}). We also collected embodiment ratings \citep{britton2024embodiment}, examined how the two sets of ratings relate to each other, and investigated whether the 11-dimensional sensorimotor construct accounts for variance beyond the single embodiment dimension in predicting lexical decision performance in Study 2.

\subsection{Participants}
A total of 378 unique participants took part in the study across two data collection rounds: 291 in the sensorimotor rating round (males = 100, females = 180, preferring not to say = 6; mean age = 31.3) and 129 in the embodiment rating round (males = 54, females = 75; mean age = 33.3). Participants were recruited primarily through \textit{Prolific} (N=312), with an additional 66 recruited via Chinese social media platforms (\textit{WeChat} and \textit{QQ}) to supplement the pool of eligible Mainland China-based participants.\footnote{The study was designed around 2023, when approximately 800 Chinese-speaking participants were available on Prolific.} On average, each participant completed 13.7 surveys in the sensorimotor rating round and 4.3 surveys in the embodiment rating round. To ensure a reasonable sample size per item, a minimum of 20 valid responses per survey was required from the participant pool. The study was approved by the Ethics Committee at The Hong Kong Polytechnic University. All participants provided informed consent and were compensated for their participation. 

\subsection{Materials}

The wordlist was assembled in two stages, yielding 3,000 frequently used stimuli in total. In the first stage, we pooled words from four existing Chinese norming studies, whose items are commonly used words that collectively span multiple grammatical categories and sensorimotor rating schemes: adjectives rated across five perceptual modalities (N = 232; \citealt{chen2019mandarin}), nouns rated across 11 sensorimotor dimensions (N = 664; \citealt{zhong2022sensorimotor}), verbs rated on embodiment and modality exclusivity (N = 689; \citealt{britton2024embodiment}), and words of mixed grammatical categories rated on a modality-general sensory experience rating (SER; N = 1,130; \citealt{wu2024sensory}). After removing duplicates, this yielded an initial set of 2,571 words. Because these studies differed in rating dimensions and collection procedures, new ratings were collected for all items under a unified 11-dimensional sensorimotor framework to ensure cross-study comparability. In the second stage, 429 additional high-frequency words absent from the existing studies were appended, selected from a weighted frequency list compiled from the Sinica Corpus 4.0 \citep{chen1996sinica} and the Traditional and Simplified Chinese parts of Chinese Gigaword 5 \citep{parker2011chinese}. 

\subsection{Procedure}

\begin{figure}[htbp]
    \centering
    \includegraphics[width=0.8\textwidth]{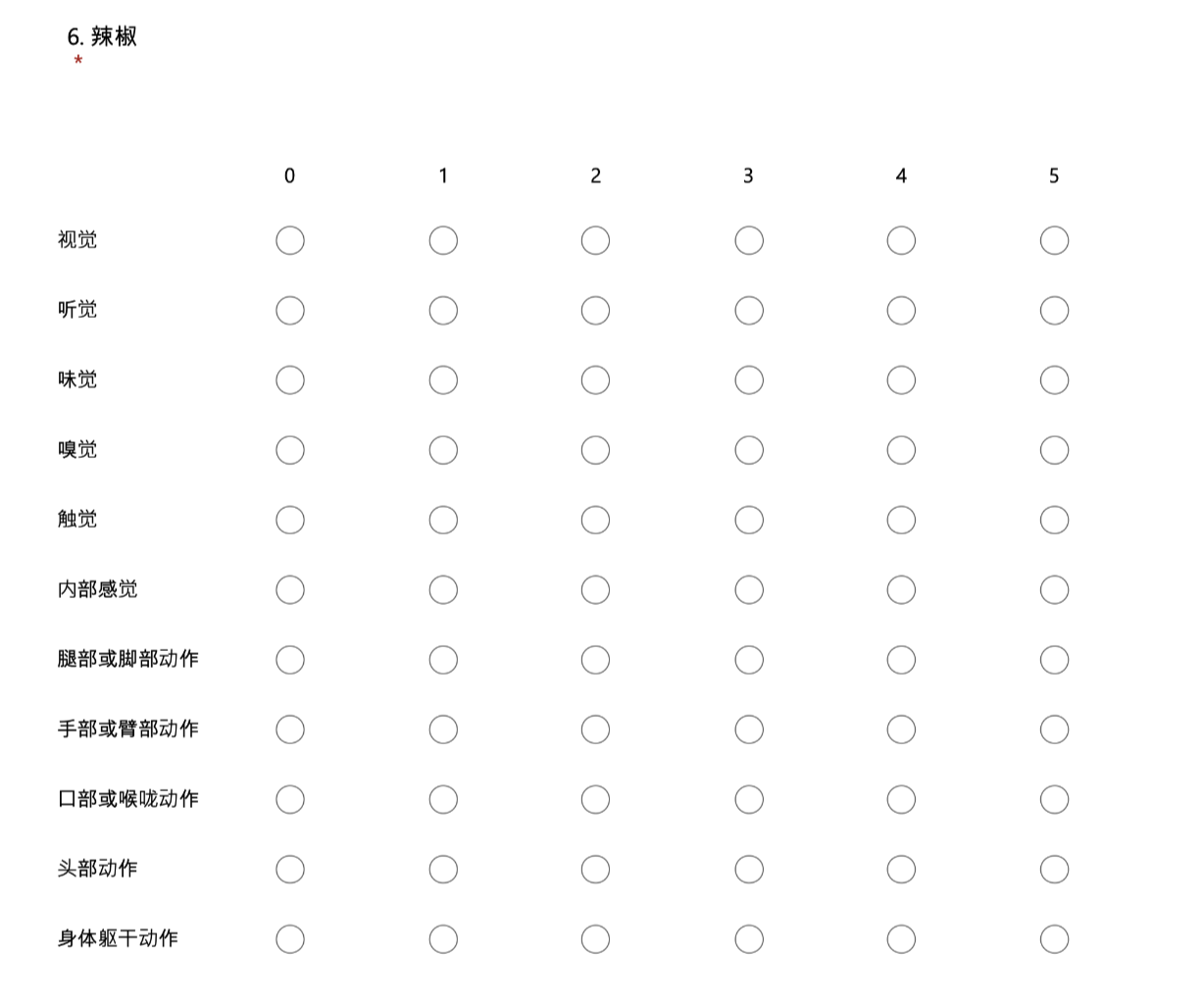}
    \caption{An example of the survey questions for a random target word: 辣椒 \textit{chili pepper}.}
    \label{fig:example}
\end{figure}

For sensorimotor ratings, items were randomly presented to participants for sensorimotor strength ratings in a norming procedure following \citet{zhong2022sensorimotor}, which in turn followed the Lancaster framework by \citet{lynott2020lancaster}. Each word was presented in a different section and was rated on 11 dimensions, including six sensory modalities (视觉 vision, 听觉 hearing, 味觉 taste, 嗅觉 smell, 触觉 touch, and 内部感觉 interoception) and five action effectors (手部或手臂部动作 hand/arm, 腿部或脚部动作 foot/leg, 头部动作 head, 口部或喉咙动作 mouth/throat, and 身体躯干动作 torso). Participants were asked to rate how strongly they felt each word on each dimension on a scale from 0 (\textit{not at all}) to 5 (\textit{greatly}). This instruction was given at the start of each survey (See Appendix \ref{sec:instructions}). An example of a trial is reported in Fig~\ref{fig:example}. Once completing the 11-dimensional ratings of each target, participants were able to proceed to the next target word. 

The embodiment ratings were collected in a separate round of surveys. Following \citet{britton2024embodiment}, participants were asked to rate how strongly each target word evoked bodily experience on a 7-point Likert scale from 0 (\textit{not at all}) to 6 (\textit{greatly}).  

In these two rounds, sensorimotor and embodiment ratings were prepared using Microsoft Forms, with each survey containing a random set of 15 target words. In addition, each survey included two dummy items serving as attention checks, such as 辣椒 \textit{chili pepper} and 油画 \textit{oil painting}, which are supposed to show higher strengths on certain dimensions than others (e.g., Gustatory $>$ Auditory, Mouth $>$ Foot/leg for \textit{chili pepper}). In addition, native speaker status was verified through a brief Chinese language screening administered at the start of each survey (See Appendix \ref{sec:mini-test}). 

\subsection{Results and discussion}

Participants who failed the Chinese language competence test or attention checks were rejected and excluded from the analysis. We examine the reliability of the ratings through commonly used Cronbach's $\alpha$ \citep{lynott2020lancaster, zhong2022sensorimotor, repetto2023italian} and validation against existing norms. We then present descriptive statistics about the mean ratings and derived metrics such as maximum strength, dominant modality, and modality exclusivity. Finally, we conduct correlation analyses to examine the relationships between different sensorimotor dimensions, and between sensorimotor and embodiment ratings.

\subsubsection{Reliability and validation}

Cronbach's $\alpha$ was computed per survey with participants treated as items, and values were averaged across surveys within each dimension. Reliability was high across all sensorimotor dimensions: Visual $= .929$, Auditory $= .909$, Gustatory $= .949$, Olfactory $= .929$, Haptic $= .881$, Interoceptive $= .823$, Hand/arm $= .831$, Mouth/throat $= .884$, Foot/leg $= .811$, Torso $= .780$, and Head $= .698$. Also, reliability was similarly high for the embodiment ratings, with a $\alpha = .885$.

To validate the norms against existing Chinese resources, we computed Spearman correlations between our ratings and those for shared stimuli. The sensorimotor ratings correlated strongly with those of \citet{zhong2022sensorimotor}, with mean $\rho = .623$ across all 11 dimensions, and with those of \citet{chen2019mandarin}, with $\rho = .716$ across the 5 perceptual dimensions. The embodiment ratings also correlated strongly with those of \citet{britton2024embodiment}, with $\rho = .821$. All correlations were statistically significant at $p < .001$.

\subsubsection{Descriptive statistics}

\paragraph{Statistics overview} 
For each concept, we computed the mean rating across all participants for each of the 11 sensorimotor dimensions and the embodiment dimension. Table~\ref{tab:descriptives} presents descriptive statistics for all dimensions. The mean rating across the 11 sensorimotor dimensions was 1.328 (range: $[0.510, 2.710]$), and the embodiment dimension had a mean of $M = 1.994$. 

\begin{figure}[htbp]
    \centering
    \includegraphics[width=\textwidth]{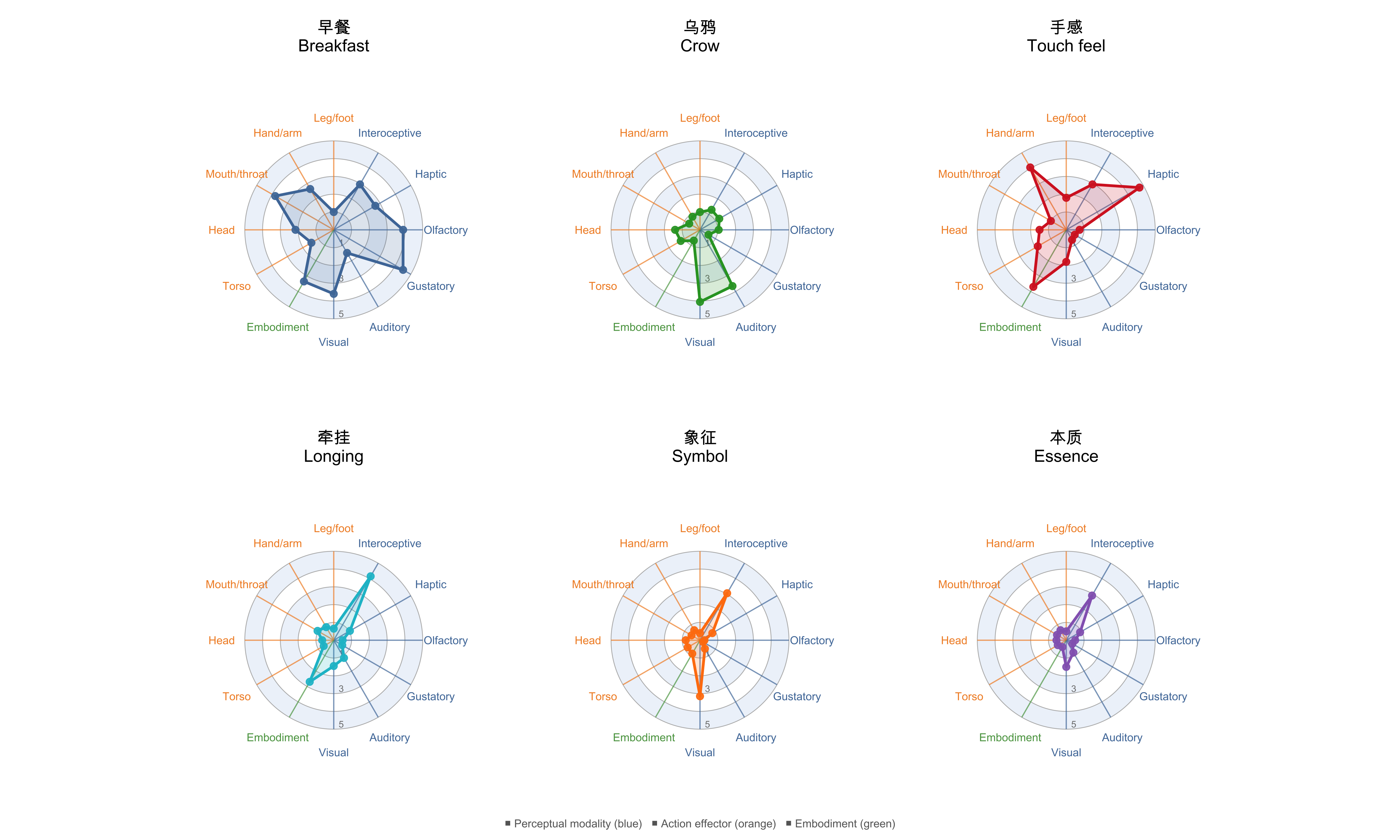}
    \caption{Sensorimotor and embodiment profiles of six example words. Each radar plot shows ratings across 11 sensorimotor dimensions and one embodiment dimension.}
    \label{fig:examples}
\end{figure}

Among perceptual modalities, Visual had the highest mean ($M = 2.710$), followed by Interoceptive ($M = 2.563$), while Gustatory ($M = 0.510$) and Olfactory ($M = 0.590$) had the lowest. Among action effectors, Hand/arm had the highest mean ($M = 1.288$), while Foot/leg had the lowest ($M = 0.937$). Figure~\ref{fig:examples} illustrates several examples of words with ratings from different dimensions.

In addition, the uniqueness values, which represent the proportion of variance in each dimension that is not shared with other dimensions, were generally high across dimensions, with Interoceptive showing the highest uniqueness ($0.836$) and Torso showing the lowest ($0.119$). 

\begin{table}[ht]
  \centering
  \caption{Descriptive statistics for 11 sensorimotor dimension and one dimension for embodiment: mean, standard deviation (SD), standard error (SE), and uniqueness. Note that the mean ratings are on a scale from 0 to 5 for sensorimotor dimensions, and from 0 to 6 for embodiment.}
  \label{tab:descriptives}
  \begin{tabularx}{\textwidth}{Xrrrr}
    \toprule
    Dimension & Mean & SD & SE & Uniqueness \\
    \midrule
    \multicolumn{5}{l}{\textit{Perceptual modality}} \\
    \quad Visual & 2.710 & 0.944 & 0.324 & 0.593 \\
    \quad Auditory & 1.617 & 0.927 & 0.316 & 0.565 \\
    \quad Gustatory & 0.510 & 0.718 & 0.173 & 0.253 \\
    \quad Olfactory & 0.590 & 0.697 & 0.187 & 0.291 \\
    \quad Haptic & 1.077 & 0.760 & 0.265 & 0.480 \\
    \quad Interoceptive & 2.563 & 0.713 & 0.343 & 0.836 \\
    \midrule
    \multicolumn{5}{l}{\textit{Action effector}} \\
    \quad Leg/foot & 0.937 & 0.646 & 0.250 & 0.225 \\
    \quad Hand/arm & 1.288 & 0.685 & 0.290 & 0.278 \\
    \quad Mouth/throat & 1.160 & 0.683 & 0.288 & 0.320 \\
    \quad Head & 1.097 & 0.507 & 0.284 & 0.246 \\
    \quad Torso & 1.064 & 0.603 & 0.270 & 0.119 \\
    \midrule
    \multicolumn{5}{l}{\textit{Embodiment}} \\
    \quad Embodiment & 1.994 & 0.920 & 0.393 & 0.627 \\
    \bottomrule
  \end{tabularx}
\end{table}

\paragraph{Dominance and exclusivity scores}

The dominant sensorimotor dimension of each concept was identified as the dimension with the highest mean rating, and the number and percentage of concepts dominant in each dimension were computed following \citet{lynott2020lancaster}. For each set of dimension-dominant concepts, we further computed mean sensorimotor strength ratings across all 11 dimensions, as well as mean exclusivity scores following \citet{lynott2009modality}, defined as the ratio of the rating range to the summed strength across all dimensions (see Equation \ref{equ:exclusivity}, where \textit{r} refers to dimensional ratings). Table~\ref{tab:dominance} presents these statistics for each dominant dimension.

\begin{equation}
  \text{Exclusivity} = \frac{\max(\mathbf{r}) - \min(\mathbf{r})}{\sum \mathbf{r}}.
\end{equation}
\label{equ:exclusivity}

\begin{figure}[htbp]
    \centering
    \begin{subfigure}[b]{0.60\textwidth}
        \centering
        \includegraphics[width=\textwidth]{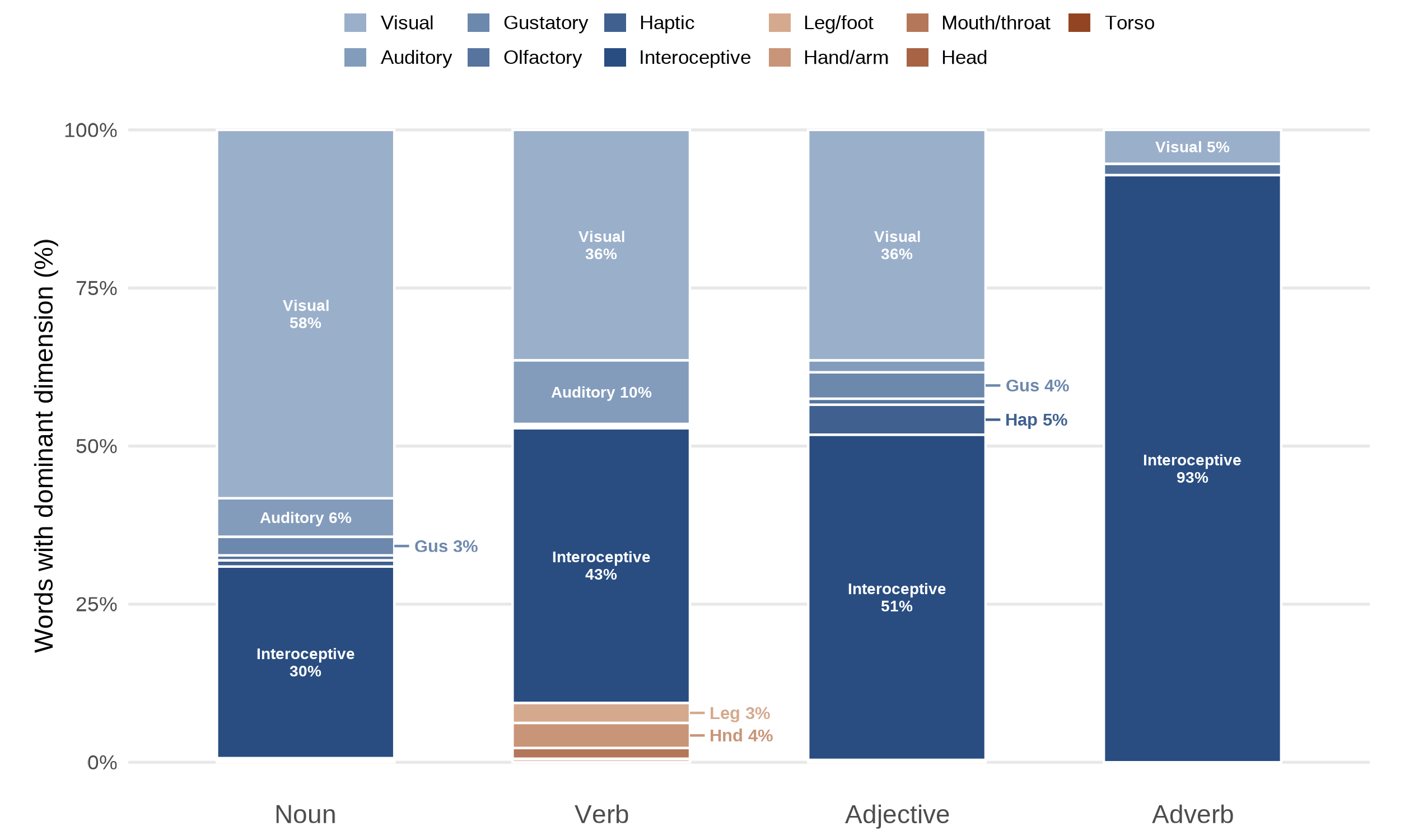}
        \caption{}
        \label{fig:pos-dominance}
    \end{subfigure}
    \hfill
    \begin{subfigure}[b]{0.35\textwidth}
        \centering
        \includegraphics[width=\textwidth]{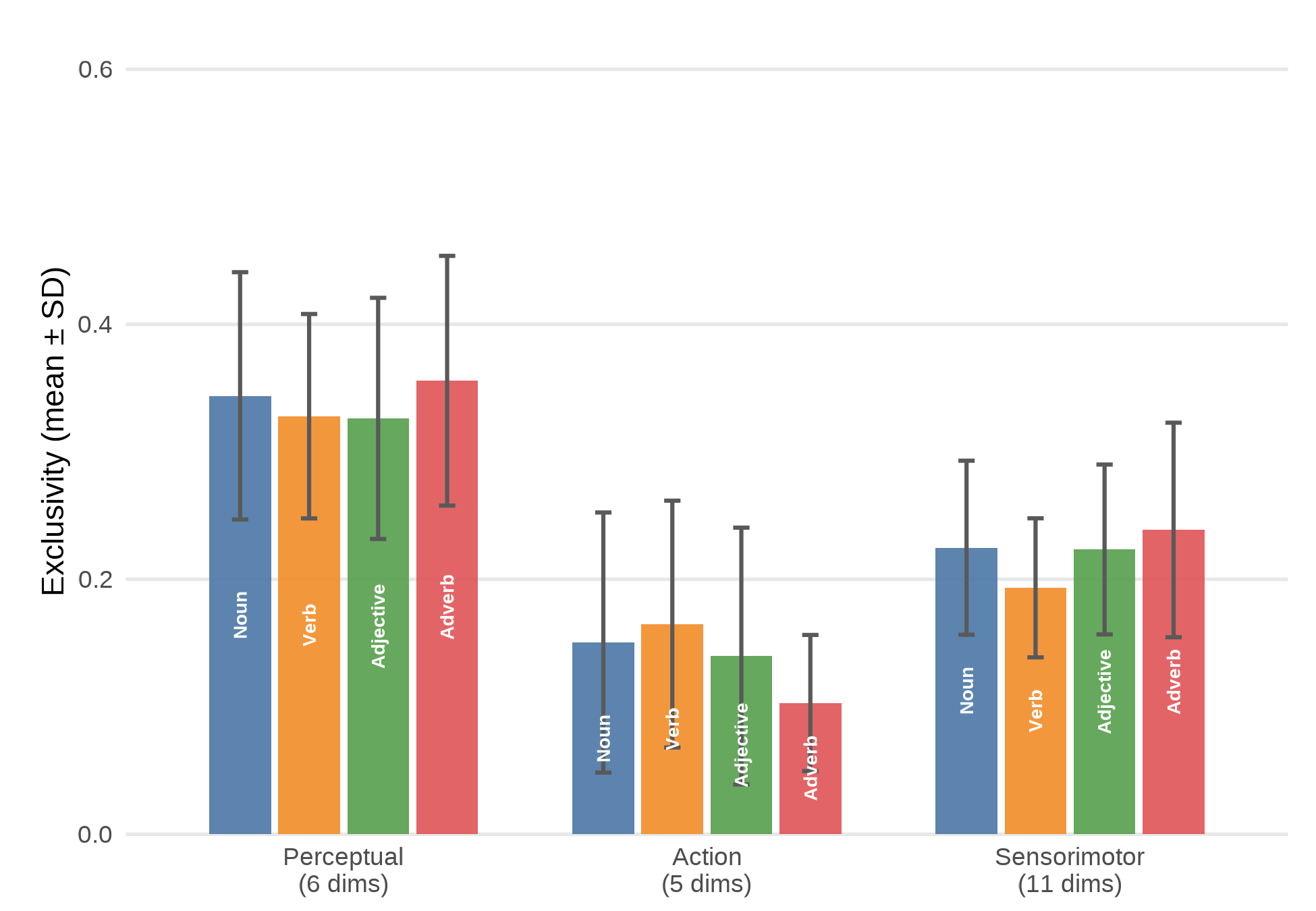}
        \caption{}
        \label{fig:pos-exclusivity}
    \end{subfigure}
    \caption{Modality dominance and exclusivity categorized by part-of-speech. (a) the proportion of words per part-of-speech category whose highest mean rating falls in each of the 11 sensorimotor dimensions; blue shades represent perceptual modality dimensions and orange shades represent action effector dimensions. (b) Mean exclusivity scores across different parts-of-speech. Nouns = 1,528; Verbs = 884; Adjectives = 527; Adverbs = 56. }
    \label{fig:pos-plots}
\end{figure}

Overall, the Chinese concepts in the database were predominantly associated with perceptual dimensions ($N = 2{,}905$, 96.8\%), with far fewer dominant in action effectors ($N = 95$, 3.2\%). Among perceptual modalities, Visual was the most common dominant dimension ($N = 1{,}409$, 47.0\%), followed by Interoceptive ($N = 1{,}173$, 39.1\%) and Auditory ($N = 192$, 6.4\%). The prevalence of visual dominance is consistent with findings from English norms \citep[e.g.,][]{winter2018vision, lynott2020lancaster} and Italian norms \citep[e.g.,][]{vergallito2020perceptual, repetto2023italian}, while the high proportion of interoceptive-dominant concepts reflects a characteristic feature of the Chinese lexical sample. Figure~\ref{fig:pos-plots} further shows the distribution of dominant dimensions and exclusivity scores by part-of-speech (POS). Perceptual dominance was observed across all POS categories (Figure~\ref{fig:pos-dominance}), particularly for visual and interoceptive, while verbs showed a higher proportion of action-effector dominance, particularly hand/arm and leg/foot. Mean strength profiles for dimension-dominant concepts are reported in Table~\ref{tab:dominance}, with chemosensory-dominant words achieving the highest mean strengths (gustatory = 3.93, olfactory = 3.89).

For exclusivity, adverbs showed the highest mean scores within the perceptual domain, while verbs led within the action-effector domain (Figure~\ref{fig:pos-exclusivity}). Across all 11 dimensions, mean exclusivity ranged from 14.5\% (Torso) to 22.5\% (Interoceptive) (Table~\ref{tab:dominance}), indicating that sensorimotor experience is generally distributed across multiple dimensions rather than concentrated in a single one. 

\begin{sidewaystable}[p]
  \centering\footnotesize
  \caption{Numbers of concepts (\textit{N} and \%) per dominant sensorimotor dimension, with mean sensorimotor strength ratings (0--5) across all 11 dimensions (Dominance) and mean exclusivity score (Exclusivity). Note: The mean rating for each dominant dimension is in bold.}
  \label{tab:dominance}
  \begin{tabularx}{\textheight}{X r r r r r r r r r r r r r r}
    \toprule
    & & & \multicolumn{11}{c}{Dominance} & \multicolumn{1}{c}{Exclusivity} \\
    \cmidrule(lr){2-14}\cmidrule(lr){15-15}
    & $N$ & \% & \rotatebox{90}{Visual} & \rotatebox{90}{Auditory} & \rotatebox{90}{Gust.} & \rotatebox{90}{Olfa.} & \rotatebox{90}{Haptic} & \rotatebox{90}{Intero.} & \rotatebox{90}{Leg/ft} & \rotatebox{90}{Hand/arm} & \rotatebox{90}{Mouth/thr.} & \rotatebox{90}{Head} & \rotatebox{90}{Torso} & Excl.\% \\
    \midrule
  \quad Visual & 1409 & 47.0\% & \textbf{3.38} & 1.58 & 0.44 & 0.63 & 1.24 & 2.25 & 1.02 & 1.39 & 1.02 & 1.14 & 1.13 & 21.2\% \\
  \quad Auditory & 192 & 6.4\% & 2.23 & \textbf{3.69} & 0.23 & 0.27 & 0.63 & 2.35 & 0.77 & 1.34 & 2.26 & 1.40 & 1.12 & 22.2\% \\
  \quad Gustatory & 70 & 2.3\% & 2.63 & 0.80 & \textbf{3.93} & 2.78 & 1.64 & 2.44 & 0.55 & 1.33 & 2.67 & 1.20 & 0.87 & 17.2\% \\
  \quad Olfactory & 19 & 0.6\% & 1.85 & 0.69 & 1.95 & \textbf{3.89} & 0.92 & 2.49 & 0.54 & 1.02 & 1.42 & 1.11 & 0.78 & 22.2\% \\
  \quad Haptic & 42 & 1.4\% & 2.69 & 1.07 & 0.57 & 0.49 & \textbf{3.71} & 2.48 & 0.81 & 1.99 & 0.88 & 0.84 & 1.04 & 20.9\% \\
  \quad Interoceptive & 1173 & 39.1\% & 2.02 & 1.42 & 0.42 & 0.43 & 0.81 & \textbf{3.03} & 0.81 & 1.04 & 1.04 & 0.96 & 0.92 & 22.5\% \\
  \midrule
  \quad Leg/foot & 29 & 1.0\% & 2.78 & 1.34 & 0.30 & 0.42 & 1.12 & 1.96 & \textbf{3.48} & 2.49 & 1.10 & 1.57 & 2.67 & 17.0\% \\
  \quad Hand/arm & 41 & 1.4\% & 2.41 & 1.34 & 0.34 & 0.35 & 1.51 & 1.92 & 1.33 & \textbf{2.85} & 1.22 & 1.33 & 1.70 & 16.3\% \\
  \quad Mouth/throat & 17 & 0.6\% & 2.19 & 2.27 & 0.91 & 0.87 & 1.01 & 2.09 & 0.62 & 1.47 & \textbf{3.09} & 1.56 & 1.05 & 17.1\% \\
  \quad Head & 1 & 0.0\% & 2.80 & 0.60 & 0.05 & 0.05 & 0.35 & 1.80 & 0.70 & 0.70 & 0.85 & \textbf{3.35} & 1.00 & 26.9\% \\
  \quad Torso & 7 & 0.2\% & 3.31 & 1.93 & 0.33 & 0.49 & 1.86 & 2.04 & 3.32 & 3.24 & 1.77 & 2.67 & \textbf{3.86} & 14.5\% \\
    \bottomrule
  \end{tabularx}
\end{sidewaystable}

\subsubsection{Correlation analyses}

\begin{figure}[htbp]
    \centering
    \includegraphics[width=\textwidth,height=0.85\textheight,keepaspectratio]{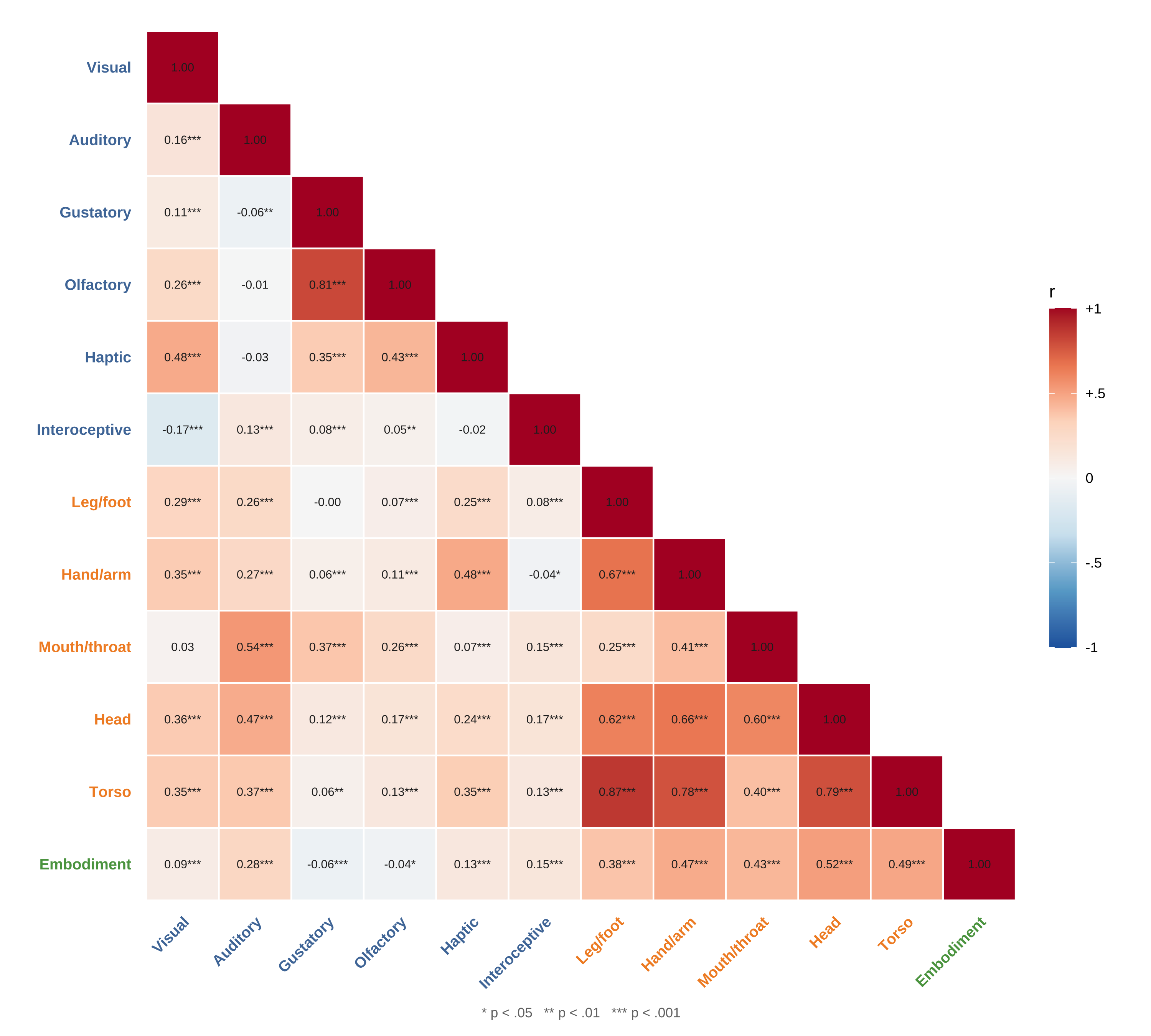}
    \caption{Correlation matrix of the 11 sensorimotor dimensions and embodiment. Color intensity represents the strength of the correlation, with blue indicating positive correlations and red indicating negative correlations.}
    \label{fig:correlation_matrix}
\end{figure}

To investigate the relations among these dimensions under sensorimotor information, we computed pairwise Pearson correlations among sensorimotor and embodiment ratings across the 3,000 concepts. The full matrix is displayed in Figure~\ref{fig:correlation_matrix}.

Several patterns replicate findings from English and Italian \citep{lynott2020lancaster,repetto2023italian}. The strongest correlation among perceptual modalities was between gustatory and olfactory strength ($r = .81$), consistent with the close co-occurrence of taste and smell in bodily experiences. Visual and haptic strength were moderately positively correlated ($r = .48$), reflecting the tendency for tangible objects to be simultaneously visible (e.g., 皮肤 \textit{skin}). The chemical senses also showed moderate positive correlations with haptic strength (olfactory–haptic: $r = .43$; gustatory–haptic: $r = .35$), suggesting that objects of smell and taste typically involve physical contact. Interoception was negatively correlated with vision ($r = -.17$), consistent with sensations arising from within the body being typically unavailable to sight (e.g., 痛 \textit{pain}, 牵挂 \textit{longing}), and was largely independent of haptic strength ($r = -.02$). Auditory was relatively orthogonal to the chemical and haptic senses (all $|r| \leq .06$), showing a small positive correlation with vision ($r = .16$) and interoception ($r = .13$), the latter potentially reflecting internally perceptible bodily sounds (e.g., 心跳 \textit{heartbeat}, auditory = 2.85, interoceptive = 3.85).

Among action effectors, ratings were strongly intercorrelated, forming a coherent limb-torso cluster: leg/foot with torso ($r = .87$), head with torso ($r = .79$), hand/arm with torso ($r = .78$), hand/arm with leg/foot ($r = .67$), and hand/arm with head ($r = .66$), reflecting the coordinated engagement of body regions in whole-body motor activity. Mouth/throat showed moderate correlations with head ($r = .60$) and hand/arm ($r = .41$), but was more weakly coupled with the distal limb effectors (leg/foot: $r = .25$), consistent with its more localised role in oral and facial action. 

We also observed notable correlations between perceptual modalities and action effectors. The strongest was between auditory and mouth/throat ($r = .54$), consistent with the oral site of sound production in auditory-dominant concepts such as 歌声 \textit{singing voice} and 口音 \textit{accent}. Haptic strength correlated positively with hand/arm ($r = .48$), reflecting the typical involvement of manual contact in tactile experience. Visual strength showed moderate associations with head ($r = .36$), hand/arm ($r = .35$), and torso ($r = .35$), plausibly reflecting the visual salience of concepts involving purposive whole-body action. The chemical senses correlated most strongly with mouth/throat (gustatory: $r = .37$; olfactory: $r = .26$), consistent with the oral cavity as the shared interface for taste and smell during ingestion. By contrast, interoceptive strength was uniformly weakly correlated with all action effectors (all $|r| \leq .17$), indicating the relative independence of internal bodily sensation from overt motor engagement.

Embodiment ratings were more closely related to action effectors (range:[$.38, .52]$) than to perceptual modalities (range:$[-.06, .28]$), with the strongest associations observed for Head ($r = .52$), Torso ($r = .49$), and Hand/arm ($r = .47$). This pattern indicates that embodiment, as a holistic bodily experience rating, primarily captures the motoric rather than perceptual dimensions of conceptual knowledge.

\section{Study 2: Sensorimotor information in lexical processing}
\label{sec:study2}

In this validation study, we examined the predictive power of sensorimotor information in lexical decision tasks. Specifically, the 11-dimension sensorimotor ratings for each concept were framed as composite variables and entered into regression models to predict accuracy and response times in lexical processing, together with the unidimensional ratings of embodiment. In addition to the six composite variables from \citet{lynott2020lancaster}, we included the two metrics derived from Perceptual Strength of Embodiment (PSE), a theoretical construct defined by both modality exclusivity and inclusivity proposed by \citet{huang2025linguistic}, which we validate on chronometric data for the first time here.

\subsection{Materials and methods}

\subsubsection{Lexical decision data and predictors}

We took 2,044 words that are present in both our sensorimotor norms and the MELD-SCH database \citep{tsang2018meld}, which provides lexical decision data for 12, 578 Chinese words from 504 native speakers. Specifically, we extracted standardized response times (\textit{zRT}) and error rates (\textit{ERR}) as dependent variables, as well as lexical predictors of \textit{word length} (in characters) and \textit{log-transformed frequencies} from SUBTLEX-CH \citep{cai2010subtlex} as control variables. 

\subsubsection{Composite sensorimotor metrics}

The sensorimotor information of concepts is inherently multidimensional, formed by 11-dimensional ratings that capture different perceptual and action-related attributes. Therefore, we derived composite variables to capture the overall sensorimotor strength of concepts in a single metric, and avoid issues of multicollinearity in regression models.

We first introduced composite variables of \textit{Perceived Strength of Embodiment} (PSE) following \citet{huang2025linguistic}, which integrates the maximum strength with modality exclusivity via the formula:
\begin{equation}
\text{PSE} = M_x + M_x(1 - E_x),    
\end{equation}
where $M_x$ is the maximum rating, $E_x$ is modality exclusivity, and $1 - E_x$ theoretically denotes modality inclusivity. This formula captures the idea that the dominant modality anchors the baseline of overall embodiment strength, while the non-dominant modalities also contribute and are scaled by the dominant rating itself. Assuming ratings range from 0 to 5, PSE ranges from 0 (all dimensions rated zero) to 10 (all dimensions rated five), with higher values indicating stronger perceived embodiment. Originally, this metric was computed over the five classical perceptual modalities. Here, we computed two variants of PSE:

\begin{enumerate}
    \item \textit{PSE-Perceptual}: a composite metric computed over the five classical perceptual senses (Visual, Auditory, Gustatory, Olfactory, Haptic) following \citet{huang2025linguistic};
    \item \textit{PSE-Sensorimotor}: an extension of PSE to all 11 sensorimotor dimensions, which additionally incorporates Interoceptive and the five action-effector dimensions.
\end{enumerate}

We also followed the approach of \citet{lynott2020lancaster} to compute composite variables, which operationalized as Minkowski distance metrics with varying parameters $m$ that determine the relative weighting of dominant dimensions, as well as a PCA-based composite. These include:

\begin{enumerate}
    \setcounter{enumi}{2}
    \item \textit{Maximum strength}: the dominant dimension of each concept across sensorimotor domains, which only considers the highest rating and ignores all other dimensions entirely;
    \item \textit{Minkowski 10} ($m = 10$): a near-maximum composite across all 11 dimensions, in which the dominant dimension disproportionately drives the score while non-dominant dimensions contribute minimally;
    \item \textit{Minkowski 3} ($m = 3$): an intermediate composite across all 11 dimensions, in which stronger dimensions receive proportionally greater weight than weaker dimensions, but all contribute substantially to the score;
    \item \textit{Euclidean distance} ($m = 2$): a balanced geometric composite across all 11 dimensions, equivalent to the Euclidean norm (vector length) of the rating vector, which gives higher ratings moderately more influence than lower ones due to squaring;
    \item \textit{Summed strength} ($m = 1$, Manhattan distance): the unweighted sum of ratings across all 11 dimensions, which is equivalent to mean strength when used as a regression predictor;
    \item \textit{PCA component}: the first principal component derived from a principal component analysis of the 11 z-standardised sensorimotor dimensions, capturing the direction of maximum variance across concepts.
\end{enumerate}

In addition, we included \textit{Embodiment ratings} to assess whether multidimensional sensorimotor composites account for unique variance beyond this holistic measure. 

\begin{enumerate}
    \setcounter{enumi}{8}
      \item \textit{Embodiment}: the unidimensional bodily experience rating collected in Study 1 \citep{britton2024embodiment}
\end{enumerate}

In total, we computed nine predictors, i.e., eight composite sensorimotor variables and the embodiment rating, which we then entered into separate regression models to predict lexical decision performance.

\subsubsection{Regression models}

Following the Lancaster Norms \citep{lynott2020lancaster}, we fitted a series of Bayesian linear regression models to predict lexical decision times and error rates. We transformed \textit{error rate} using \textit{arcsine-square-root} for distribution normalization. For each of the nine metrics, we compared a \textit{null model} containing only \textit{word length} and \textit{frequency} against an \textit{alternative model} that additionally included a single composite variable. Models were fitted using the \texttt{BayesFactor} package in R, with a Jeffreys--Zellner--Siow (JZS) prior using the default medium scale ($r = 0.354$), identical to the JASP defaults used in \citet{lynott2020lancaster}.

\subsection{Results and discussion}

Table \ref{tab:bf_regression} presents the results of Bayesian linear regression models predicting lexical decision performance with and without composite variables for sensorimotor information. All alternative models provided strong evidence over the null model containing word length and frequency \citep{kass1995bayes}, with Bayes Factors ranging from $\ln(\text{BF}_{10}) = 7.39$ to $15.48$ for response times and from $10.17$ to $25.82$ for error rates. Critically, all regression coefficients were negative, indicating the facilitatory effects of sensorimotor information in lexical processing. Also, the incremental variance explained by alternative models ($\Delta R^2$ ranging from .006 to .023) demonstrates that sensorimotor information contributes unique predictive power beyond the controlling variables.

Among composite variables, \textit{PSE-Sensorimotor} and \textit{Minkowski-3} achieved the best performance: PSE-Sensorimotor ranked first for response times ($\ln(\text{BF}_{10}) = 15.48$, $\Delta R^2 = .011$) and second for error rates ($\ln(\text{BF}_{10}) = 23.51$, $\Delta R^2 = .021$), while Minkowski-3 ranked first for error rates ($\ln(\text{BF}_{10}) = 25.82$, $\Delta R^2 = .023$) and second for response times ($\ln(\text{BF}_{10}) = 14.86$, $\Delta R^2 = .010$). Both outperformed the undifferentiated-weighting \textit{Summed strength} and the single-dimension \textit{Maximum strength}, consistent with their shared principle of differentially up-weighting the dominant sensorimotor dimension while retaining contributions from non-dominant ones. Furthermore, \textit{PSE-Perceptual} underperformed \textit{PSE-Sensorimotor} in both scenarios, with the gap more pronounced for response times ($\ln(\text{BF}_{10}) = 8.37$ vs.\ $15.48$) than for error rates ($\ln(\text{BF}_{10}) = 21.07$ vs.\ $23.51$), indicating that extending the PSE metric to incorporate interoception and action-related dimensions yields substantial additional predictive value beyond the five classical perceptual modalities. \textit{Embodiment} ratings ranked among the weakest predictors overall, suggesting that the multidimensional structure of sensorimotor composites captures richer information than this unidimensional measure, which was found to correlate most strongly with action effectors in Study 1.

The competitive predictive performance of Minkowski-3 replicates findings from the Lancaster Norms \citep{lynott2020lancaster}, providing cross-linguistic evidence of its robustness. PSE, originally proposed to address the directionality prediction dilemma in linguistic synesthesia \citep{huang2025linguistic}, shares this emphasis on the dominant dimension but operationalizes it differently. Motivated by the observation that most sensory words are multimodal \citep{lynott2013modality, chen2019mandarin}, PSE treats the dominant modality as an anchor while scaling the supplementary contribution of non-dominant modalities by modality inclusivity ($1 - E_x$). Consequently, concepts with strength distributed across multiple dimensions (i.e., high inclusivity) receive a greater boost from non-dominant modalities, yielding a higher overall embodiment score than concepts of equivalent dominant strength that are more unimodal. 

In summary, composite variables that disproportionately weight the dominant dimension while retaining contributions from the remaining dimensions provide a more informative and predictive representation of sensorimotor information for lexical decision performance.

\begin{table}[ht]
\centering
\caption{Bayesian linear regression results for lexical decision performance prediction. The null model contains word length and frequency; alternative models additionally include one composite variable. Models were fitted with a JZS prior ($r = 0.354$). $\ln(\text{BF}_{10})$: natural log Bayes factor (alternative vs.\ null); $\beta$: unstandardised regression coefficient for composite variables; $t$: $t$-statistic; $R^2$: variance explained by the model; $\Delta R^2$: incremental variance explained over the null model. All $p < .0001$.}
\label{tab:bf_regression}
\small
\setlength{\tabcolsep}{4pt}
\begin{tabular*}{\textwidth}{@{\extracolsep{\fill}}llrrrrr@{}}
\toprule
\multicolumn{7}{c}{\textbf{\textit{zRT}}} \\
\midrule
 & Composite variable & $\ln(\text{BF}_{10})$ & $\beta$ & $t$ & $R^2$ & $\Delta R^2$ \\
\midrule
\textit{Null model} &  &    &     &      & \textbf{.410} &   \\
\addlinespace[0.3em]
 & \textbf{PSE-Sensorimotor}  & \textbf{15.48} & $-$0.023 & $-$6.14 & .421 & \textbf{.011} \\
 & Minkowski-3       & 14.86 & $-$0.030 & $-$6.04 & .420 & .010 \\
 & Minkowski-10      & 14.54 & $-$0.040 & $-$5.99 & .420 & .010 \\
 & Maximum strength  & 13.12 & $-$0.038 & $-$5.74 & .419 & .009 \\
 & Euclidean         & 13.05 & $-$0.019 & $-$5.73 & .419 & .009 \\
 & Summed strength   & 10.24 & $-$0.005 & $-$5.20 & .418 & .008 \\
 & PCA component     &  8.51 & $-$0.011 & $-$4.85 & .417 & .007 \\
 & PSE-Perceptual    &  8.37 & $-$0.015 & $-$4.82 & .417 & .007 \\
 & Embodiment        &  7.39 & $-$0.024 & $-$4.61 & .416 & .006 \\
\midrule
\addlinespace[0.5em]
\multicolumn{7}{c}{\textbf{\textit{ERR}$_{\text{asin}}$}} \\
\midrule
\textit{Null model} &  &    &     &      & \textbf{.162} &   \\
\addlinespace[0.3em]
 & \textbf{Minkowski-3}       & \textbf{25.82} & $-$0.022 & $-$7.62 & .185 & \textbf{.023} \\
 & PSE-Sensorimotor  & 23.51 & $-$0.016 & $-$7.30 & .183 & .021 \\
 & Minkowski-10      & 23.14 & $-$0.028 & $-$7.25 & .183 & .021 \\
 & Euclidean         & 22.07 & $-$0.014 & $-$7.09 & .182 & .020 \\
 & PSE-Perceptual    & 21.07 & $-$0.013 & $-$6.94 & .181 & .019 \\
 & Maximum strength  & 20.93 & $-$0.027 & $-$6.92 & .181 & .019 \\
 & Summed strength   & 15.10 & $-$0.004 & $-$5.99 & .176 & .015 \\
 & PCA component     & 12.25 & $-$0.007 & $-$5.48 & .174 & .012 \\
 & Embodiment        & 10.17 & $-$0.015 & $-$5.08 & .172 & .010 \\
\bottomrule
\end{tabular*}
\end{table}

\section{Study 3: How much sensorimotor information can be predicted from linguistic representations?}

In this study, we investigated to what extent sensorimotor information is recoverable from linguistic representations, adding new empirical data to the debate on whether language models as disembodied systems can acquire sensorimotor information \citep{xu2025large}. Language models are trained on pure language data, and their concept representations are built from distributional co-occurrence patterns within a self-contained symbolic system, without direct grounding in everyday experience \citep{connell2024can,contreras2025distributional}. Language, however, serves as the primary medium of human communication and encodes embodied relations \citep{louwerse2008embodied} and in turn serves as a source of grounding \citep{dove2023rethinking, dove2023language}. Therefore, if sensorimotor information is indeed encoded in language use, it should be recoverable from linguistic representations such as word embeddings, which are trained on large corpora and capture statistical patterns of word usage \citep{lenci2008distributional, gunther2019vector}.

Specifically, we tested the mapping between linguistic representation and sensorimotor ratings by fitting regression models on a training set and evaluating on held-out data \citep{chersoni-etal-2020-automatic}. We then quantified (1) the alignment between predicted and human-rated norms for each dimension, and (2) whether the predicted sensorimotor geometry mirrors the structure of the human-rated sensorimotor space. We used relatively simple regression models on static word embeddings rather than large language models, avoiding confounds such as multimodal training data and data contamination.

\subsection{Materials and methods}

\subsubsection{Linguistic representations and shared stimulus set}

We obtained the off-the-shelf representations from the Tencent AI Lab Chinese word embeddings \citep{song-etal-2018-directional}, a large-scale model trained on Chinese web pages and \textit{Tencent News} covering eight million Chinese words and phrases.\footnote{\url{https://modelscope.cn/models/lili666/text2vec-word2vec-tencent-chinese/summary}} This model was selected for its broad vocabulary coverage of Mandarin Chinese. Of the 3,000 concepts in our sensorimotor norms, embeddings were available for 2,859 (95.3\%), which constituted the shared stimulus set for all subsequent analyses.

\subsubsection{Ridge Regression}

For each of the 11 sensorimotor dimensions, we fitted a ridge regression model that linearly maps the word embeddings trained from pure textual data to a predicted rating representing experiential strength from that dimension. The regularization strength ($\alpha$) was selected via cross-validation within the training data. To obtain unbiased predictions for all 2,859 concepts, we used 10-fold cross-validation: the concepts were split into 10 equal partitions, and in each iteration the model was trained on nine partitions and used to predict ratings for the remaining one, so that every concept received exactly one out-of-fold predicted rating. All models were implemented in the \textit{scikit-learn} Python package \citep{pedregosa2011scikit}.

Model performance was quantified using two complementary metrics: the coefficient of determination ($R^2$, ranging from 0 to 1), which captures the proportion of variance in ratings accounted for by the model, and Spearman $\rho$ between observed and predicted ratings \citep{chersoni-etal-2020-automatic}, which is robust to outliers and non-linear scaling. Higher values indicate better predictive accuracy for both metrics. 

To assess whether observed performance exceeded chance, significance was evaluated via a one-sided permutation test with 1{,}000 iterations, in which word-rating labels were randomly shuffled to generate a null distribution, and $p$-values were corrected via false discovery rates across 11 dimensions. In addition, 95\% bootstrap confidence intervals (CIs) were computed from 1{,}000 bootstrap repetitions, each drawing a random sample of 1{,}000 words with replacement from the full set of 2{,}859 concepts.

\subsubsection{Representational Similarity Analysis (RSA)}

We used RSA to test whether embedding-based predictions preserve the geometric structure of the true sensorimotor space. For each of the 2,859 concepts, we constructed an 11-dimensional vector by z-scoring each dimension independently, yielding a true sensorimotor matrix (from human ratings) and a predicted sensorimotor matrix (from out-of-fold ridge regression predictions). Pairwise Euclidean distances between these vectors yielded two $2{,}859 \times 2{,}859$ RDMs. Their alignment was quantified as the Spearman correlation between the upper triangles of the two RDMs \citep{kriegeskorte2013representational}, with higher values indicating greater structural correspondence between the predicted and true sensorimotor spaces. Significance was evaluated via a one-sided permutation test with 500 iterations, in which the word labels of the predicted RDM were randomly shuffled by permuting its rows and columns simultaneously to generate a null distribution.

\subsection{Results and discussion}

\begin{figure}[htbp]
    \centering
    \includegraphics[width=\textwidth]{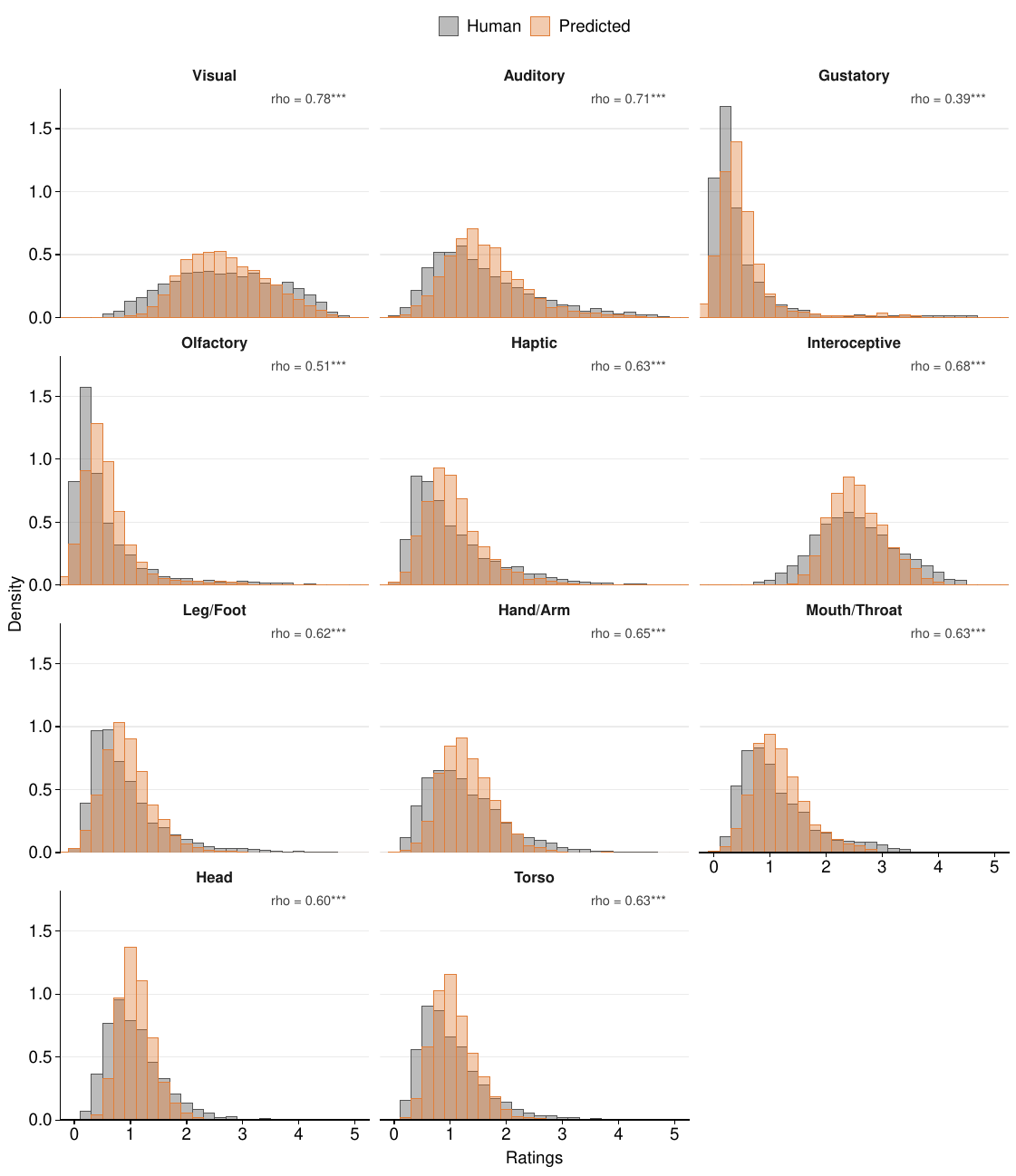}
    \caption{Distribution of human and predicted sensorimotor ratings across 11 dimensions. Each panel shows the density histogram of ratings on a 0–5 scale for one sensorimotor dimension. Grey bars represent human generated ratings, while orange bars represent ratings predicted by the ridge regression model. Spearman $\rho$ between observed and predicted ratings is shown in the top right corner of each panel. All correlations were significant at $p < .001$.}
    \label{fig:human_predicted_ratings}
\end{figure}

Figure~\ref{fig:human_predicted_ratings} shows the rating distributions across all 11 dimensions for human norms and model predictions, with predicted distributions narrower than their counterparts. All dimensions yielded significant Spearman correlations with human ratings (mean $\rho$ = .62; all corrected $p < .001$, permutation test, $n = 1{,}000$ permutations, FDR-corrected), with $\rho$ ranging from $.39$ (Gustatory) to $.78$ (Visual). Table~\ref{tab:human_predicted_ratings} shows the human and predicted ratings for two example words. 

\begin{table}[ht]
\centering
\caption{Human and predicted sensorimotor ratings for two example words.}
\begin{tabular*}{\textwidth}{@{\extracolsep{\fill}}lcccc@{}}
\toprule
 & \multicolumn{2}{c}{\textbf{早餐} (breakfast)} & \multicolumn{2}{c}{\textbf{牵挂} (longing)} \\
\cmidrule(lr){2-3} \cmidrule(lr){4-5}
\textbf{Dimension} & \textbf{Human} & \textbf{Predicted} & \textbf{Human} & \textbf{Predicted} \\
\midrule
Visual        & 3.600 & 3.264 & 1.450 & 2.100 \\
Auditory      & 1.500 & 1.176 & 1.150 & 1.477 \\
Gustatory     & 4.500 & 3.042 & 0.550 & 0.484 \\
Olfactory     & 3.900 & 2.310 & 0.500 & 0.349 \\
Haptic        & 2.700 & 1.550 & 1.050 & 0.892 \\
Interoceptive & 2.950 & 2.374 & 4.150 & 3.607 \\
Leg/Foot      & 1.000 & 1.037 & 0.650 & 0.583 \\
Hand/Arm      & 2.650 & 1.904 & 0.850 & 0.984 \\
Mouth/Throat  & 3.800 & 2.584 & 1.050 & 0.997 \\
Head          & 2.150 & 1.604 & 0.650 & 0.920 \\
Torso         & 1.450 & 1.348 & 0.650 & 0.800 \\
\bottomrule
\end{tabular*}
\label{tab:human_predicted_ratings}
\end{table}

\begin{figure}[htbp]
    \centering
    \includegraphics[width=0.7\textwidth]{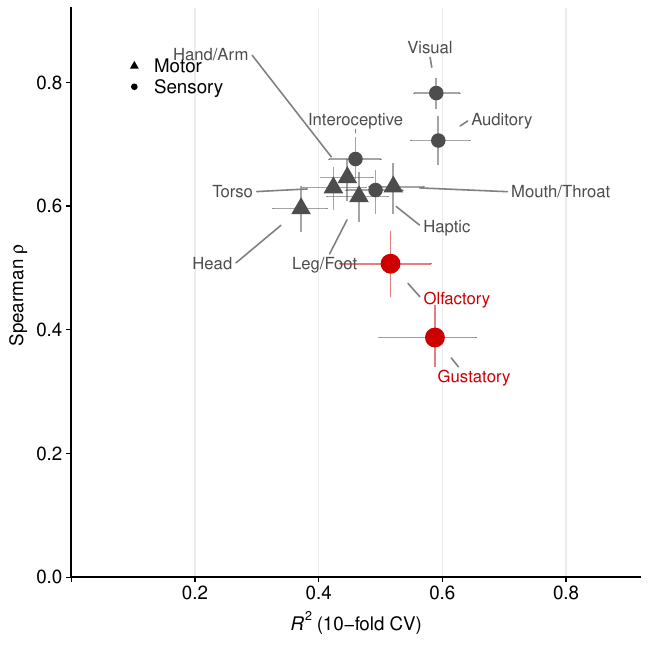}
    \caption{Relationship between cross-validated $R^2$ and Spearman $\rho$ across 11 sensorimotor dimensions. Each point represents one dimension: circles denote sensory dimensions and triangles for motor. Error bars show 95\% bootstrap confidence intervals.}
    \label{fig:scatter_r2_rho}
\end{figure}

Figure~\ref{fig:scatter_r2_rho} plots $R^2$ and Spearman $\rho$ for each dimension. These two diverge most for Gustatory and Olfactory, which show comparatively high $R^2$ alongside the lowest $\rho$ values, whereas Visual and Auditory show a more consistent relationship between the two metrics, with high $R^2$ and $\rho$. For Gustatory and Olfactory, the combination of high $R^2$ alongside low $\rho$ is consistent with heavy ridge shrinkage driven by a weak embedding-to-rating signal: when the embeddings contain little variance linearly predictive of experiential ratings, the regularization penalty compresses predictions toward the mean, collapsing rank distinctions among individual words (depressing $\rho$) while leaving the global linear trend less affected (preserving $R^2$). This weak signal is itself partly a consequence of highly skewed rating distributions on these dimensions (over 80\% of words rated below 1.0, see Gustatory and Olfactory in Fig.~\ref{fig:human_predicted_ratings}), which compress usable target variance and inherently limit the rank structure recoverable from distributional representations. Visual and Auditory, by contrast, show high and consistent values on both metrics, indicating that word co-occurrence patterns carry a strong and consistent signal for these dimensions. 

\begin{figure}
    \centering
    \includegraphics[width=\textwidth]{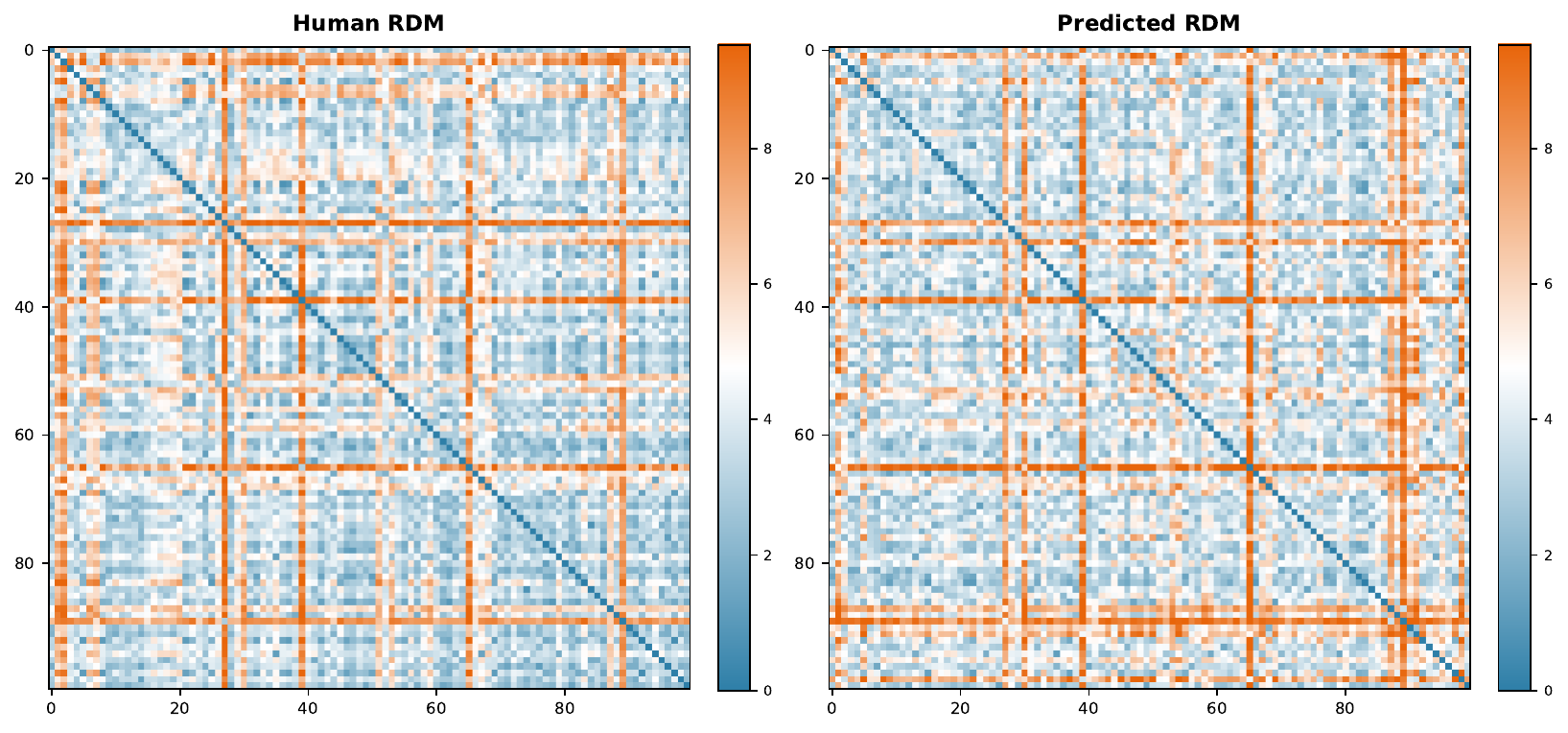}
    \caption{Representational dissimilarity matrices (RDMs) for human sensorimotor ratings (left) and ridge-regression-predicted ratings (right), visualized on a random subset of 100 words. Each cell represents the pairwise Euclidean distance between two words in the 11-dimensional sensorimotor space (z-scored per dimension). Color scale runs from blue (low dissimilarity) through white to orange (high dissimilarity), }
    \label{fig:rdm}
\end{figure}

Beyond dimension-by-dimension predictability, we examined the structural correspondence between the two sensorimotor spaces. Across the full set of words, the correlation between the two RDMs was quantified through Spearman rank correlation across all word pairs ($N = 2{,}859$; RSA), with $\rho = .540$ (95\% CI $[.514, .569]$, $p < .001$, permutation test). This indicates that the relational structure of sensorimotor space is also recoverable from linguistic representations. Figure~\ref{fig:rdm} shows the RDMs derived from human ratings and predicted ratings on a random subset of 100 words for clarity. 

Taken together, linguistic representations, reflecting statistical patterns of language use, can predict sensorimotor information, but the extent of recovery varies across dimensions. For example, vision information is richly encoded in language \citep{winter2018vision}, and linguistic representations provide a closer approximation of sensorimotor knowledge for visual dimensions compared to chemosensory dimensions such as gustatory and olfactory, which have higher language ineffability \citep{majid2018differential, kurfali2025representations}.

\section{Conclusions}

We introduced 11-dimensional sensorimotor and unidimensional embodiment norms for 3,000 lexicalized concepts in Mandarin Chinese, demonstrated their reliability and cross-norm validity, and used them to address two further questions: how to optimally aggregate multidimensional sensorimotor information for behavioral prediction, and how much of that information is recoverable from linguistic representations.

Consistent with the dominance of vision reported in other language norms \citep{winter2018vision, lynott2020lancaster, repetto2023italian, speed2022dutch}, visual strength dominates the sensorimotor space in our norms, with 47.0\% of concepts rated highest on this dimension. By contrast, chemosensory dimensions are the least differentiated, with the lowest mean ratings overall, yet concepts dominant in these dimensions achieve the highest mean strengths (see Table~\ref{tab:dominance}). Our norms also show a notably high proportion of interoception-dominant concepts (39.1\%), which occurs prominently among adjectives, adverbs, and nouns denoting abstract concepts \citep{zhong2022sensorimotor} (Figure~\ref{fig:pos-dominance}). The correlation structure of the dimensions (Figure~\ref{fig:correlation_matrix}) reveals systematic relations across the sensorimotor space, with the notable exception of interoceptive strength, which is negatively correlated with visual ($r = -.17$) and largely independent of haptic ($r = -.02$) strength, reflecting the characteristic inaccessibility of internal bodily sensations to sight and touch. Embodiment ratings correlated more strongly with action effectors than with perceptual modalities, suggesting that holistic bodily experience is more closely tied to motor engagement than to perceptual strength.

Given the multidimensionality of sensorimotor information, how to best frame it as a single composite variable for regression analyses is a critical question. Our findings suggest that composite variables spanning all 11 dimensions tend to provide richer signal for lexical decision performance, especially those that disproportionately weight the dominant dimension while retaining contributions from non-dominant dimensions (e.g., Minkowski-3 and PSE-Sensorimotor), compared to undifferentiated composites (e.g., Summed strength) or single-dimension metrics (e.g., Maximum strength). In addition to the robustness of the Minkowski-3 metric, PSE-Sensorimotor is more theoretically transparent than Minkowski composites: the contribution of each dimension is explicitly governed by modality inclusivity, making the resulting score interpretable in terms of the underlying perceptual and motor structure of each concept \citep{huang2025linguistic}.

Beyond behavioral prediction, ridge regression models fitted on static word embeddings derived from purely textual data showed robust predictive power for sensorimotor ratings across dimensions, with dimensions more richly encoded in language (e.g., vision and audition) yielding stronger recovery than those with higher language ineffability (e.g., gustation and olfaction). This asymmetry partly reflects the highly compressed rating distributions of chemosensory dimensions: over 80\% of concepts receive near-zero ratings, which limits the usable variance that distributional representations can exploit \citep{majid2018differential, kurfali2025representations}. The RSA correlation ($\rho = .540$) between predicted and human-rated sensorimotor spaces further indicates that the relational geometry of the sensorimotor space is also partially recoverable from linguistic representations, extending dimension-level recovery to the level of inter-concept structure. These results are consistent with the view that distributional patterns in language partially reflect embodied conceptual structure \citep{louwerse2008embodied}, and with proposals that language can serve as a source of sensorimotor grounding \citep{dove2023rethinking, dove2023language}.

Together, these findings provide a novel normative resource for embodied cognition research in Mandarin Chinese and offer new empirical traction on how sensorimotor structure relates to the distributional properties of language.

\section*{Appendix}\label{sec:appendix}

\subsection{Survey instructions}\label{sec:instructions}

\noindent\textbf{{\large 问卷填写说明 | INSTRUCTIONS}}

\medskip

视觉让我们欣赏这个五彩缤纷的世界；听觉让我们能够聆听各种声音的旋律；味觉让我们品尝到酸甜苦辣的味道；嗅觉让我们嗅到芬芳或臭气扑鼻的气味；触觉让我们感知到物体的温度、纹理以及身体的感觉。而内部感觉让我们察觉到饥饿、疲惫、厌恶等感受。

此外，我们还通过身体部位的移动来完善这些感知体验。例如，当我们观察周围环境时，我们需要转动头部或身体躯干；当我们说话时，我们需要张开嘴巴并利用喉咙发声；我们用手（和手臂）去触摸和抓取物体；我们还会利用脚（和腿）来行走或踢东西。

我们邀请您从上述提到的多个感知维度来对给定的目标词语进行打分。评分范围从0到5。0表示没有感觉，5表示非常强烈的感觉。数字越大，表示您越强烈地感受到目标词与该感官类别的关联性。请注意，在这个任务中没有对或错的答案。请放心提供您的主观意见。

\medskip

The visual sense allows us to see this beautiful world; the auditory sense permits us to hear sounds; the gustatory sense considers tastes; the olfactory sense accounts for smells and odors; the tactile sense perceives temperature, pain, and textures of objects; and the interoceptive sense detects the feelings of hunger, exhaustion, disgust, and other internal signals.

We also use our body parts to complete these perceptions. For example, we may move our head or torso when we view the surroundings, we open our mouth and articulate with our throat when we talk, we use our hands (and arms) to reach and grasp for objects, and we also resort to our feet (and legs) to walk or to kick.

In what follows, you will see some words in Chinese. We would like you to judge how much you experience these concepts through each of the perceptions and actions we ask, on a scale from 0 = no feelings at all, to 5 = very strong feelings. The scale that you will be using for your judgments ranges from 0 to 5. The larger the number, the stronger you feel that the target word is related to the category. Note that there are no right or wrong answers to the task, please provide us your subjective opinion.

\subsection{Mini-test on Chinese Language}\label{sec:mini-test}

Both surveys included a four-question mini-test at the outset to verify that participants were competent Mandarin Chinese users. The tests followed the same structure across surveys but used different items; participants who failed two or more questions were excluded from analysis.

\subsubsection*{Sensorimotor survey mini-test}

\begin{enumerate}
    \item \textbf{[请根据拼音写汉字 ``tiānqì qínglǎng'']} Participants wrote the Chinese characters for the pinyin phrase ``tiānqì qínglǎng'' (expected: 天气晴朗 or its traditional form, `the weather is fine'). This probed basic literacy.
    \item \textbf{[汉字 ``清'' 的偏旁部首是]} Participants identified the radical of the character 清 (expected: 氵). This tested knowledge of basic character structure.
    \item \textbf{[请选择字形正确的词]} Participants selected the correctly written word from four options (追捧; 历害; 伸吟; 即然; expected: 追捧), where three options contained ill-formed characters. This assessed familiarity with Chinese orthography and was the most difficult item in the set.
    \item \textbf{[白. 请选出下列不是 ``白'' 所表达意义的一项]} Participants identified which option is NOT a meaning of 白 \textit{bái} `white' from four options (表示很多 `indicates a lot'; 没有效果 `ineffective'; 像霜或雪的颜色 `the color of frost or snow'; 用白眼珠看人，表示轻视或不满 `looking at someone with the whites of one's eyes, indicating contempt'; expected: 表示很多). This tested culturally grounded semantic knowledge.
\end{enumerate}

\subsubsection*{Embodiment survey mini-test}

The embodiment survey mini-test used the same four question types but with different items:

\begin{enumerate}
    \item \textbf{[请根据拼音写汉字 ``yángguāng cànlàn'']} Participants wrote the Chinese characters for the pinyin phrase ``yángguāng cànlàn'' (expected: 阳光灿烂 or its traditional form, `sunny and bright'). This probed basic literacy.
    \item \textbf{[汉字 ``港'' 的偏旁部首是]} Participants identified the radical of the character 港 \textit{gǎng} `harbor' (expected: 氵). This tested knowledge of basic character structure.
    \item \textbf{[请选择字形正确的词]} Participants selected the correctly written word from four options (教室; 教寿; 教帅; 教沙; expected: 教室), where three options contained ill-formed characters. This assessed familiarity with Chinese orthography.
    \item \textbf{[请选出下列不是 ``高'' 所表达意义的一项]} Participants identified which option is NOT a meaning of 高 \textit{gāo} `high/tall' from four options (声音; 三角形等从底部到顶部的垂直距离; 高度; expected: 声音). This tested culturally grounded semantic knowledge.
\end{enumerate}

\bibliography{references.bib}

\end{CJK}
\end{document}